\RequirePackage{fix-cm}
\documentclass[twocolumn]{svjour3}          
\smartqed  
\usepackage{times}
\usepackage{epsfig}
\usepackage{graphicx}
\usepackage{amsmath}
\usepackage{amssymb}
\usepackage{epstopdf} 
\usepackage{marvosym}
\begin{document}

\title{Detecting Ground Control Points via Convolutional Neural Network for Stereo Matching
}


\author{Zhun Zhong       \and
        Songzhi Su  \and
        Donglin Cao  \and
        Shaozi Li  
}


\institute{ Zhun Zhong       \and
        Songzhi Su  \and
        Donglin Cao  \and
        Shaozi Li (\Letter) \at
			Cognitive Science Department, Xiamen University, \\ Xiamen, 361005, Fujian, China \\
              \email{zhunzhong@stu.xmu.edu.cn}           
           \and
           Shaozi Li \at
              \email{szlig@xmu.edu.cn} \and         
}

\date{Received: date / Accepted: date}

\maketitle

\begin{abstract}
   In this paper, we present a novel approach to detect ground control points (GCPs) for stereo matching problem.
   First of all, we train a convolutional neural network (CNN) on a large stereo set, and compute the matching confidence of each pixel by using the trained CNN model. 
   Secondly, we present a  ground control points selection scheme according to the maximum matching confidence of each pixel. 
   Finally, the selected GCPs are used to refine the matching costs, and we apply the new matching costs to perform optimization with semi-global matching algorithm for improving the final disparity maps. 
   We evaluate our approach on the KITTI 2012 stereo benchmark dataset. Our experiments show that the proposed approach significantly improves the accuracy of disparity maps.
 
\keywords{Stereo Maching \and CNN \and GCPs \and Maching Confidence}
\end{abstract}

\section{Introduction}

   Stereo matching is one of the most extensively researched topics in the study of computer vision. 
   The depth information computed by stereo matching algorithm can be used in various vision applications, such as 3D reconstruction, object recognition, object tracking, and autonomous navigation. 
   Scharstein and Szeliski\cite{scharstein2002taxonomy} developed a taxonomy dividing the stereo matching algorithms into four steps:
   
   (1) matching cost computation;
   
   (2) cost aggregation;
   
   (3) disparity computation/optimization; and
   
   (4) disparity refinement.
   
\begin{figure}[t]
\begin{center}
   \includegraphics[width=1.0\linewidth]{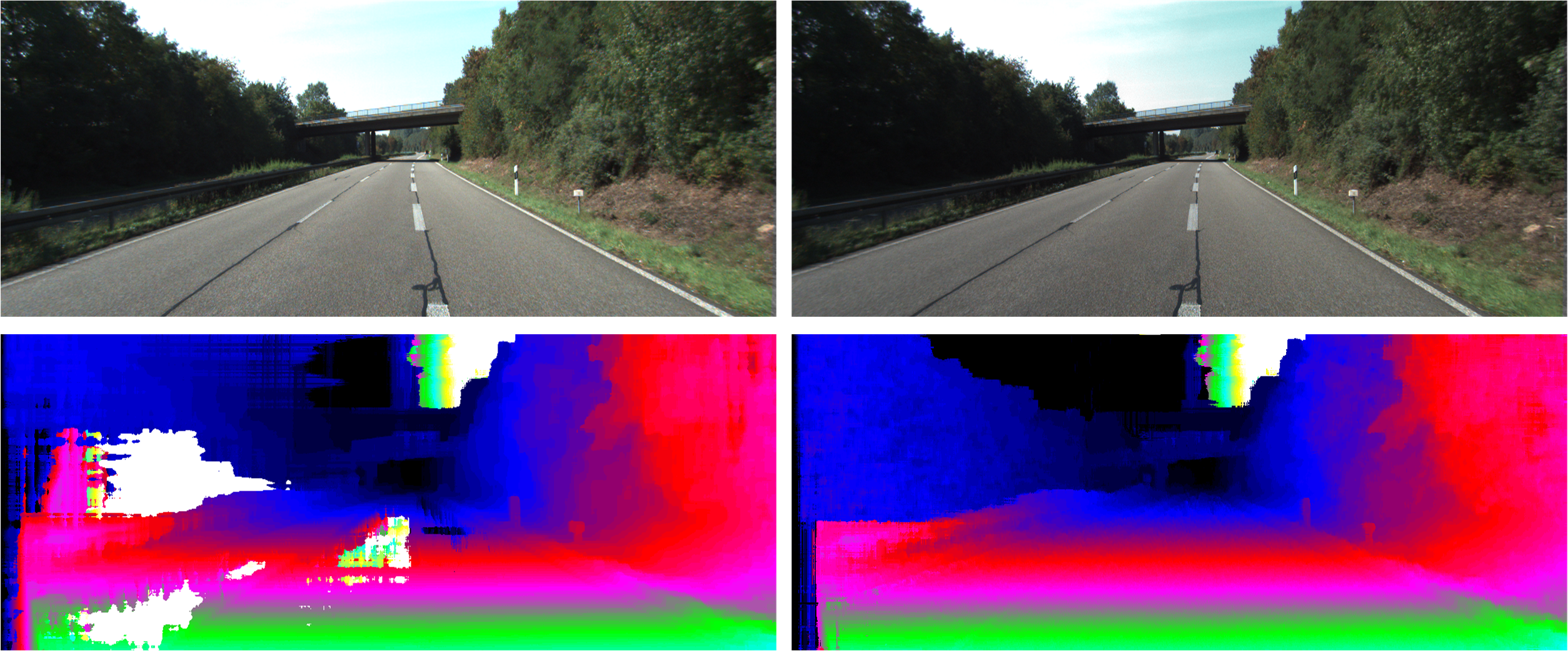}
\end{center}
   \caption{Example of our algorithm result from KITII 2012 frame.  \textbf{The top row}: left input image and right input image; \textbf{The buttom row}: disparity map of left image using SAD matching costs with SGM, and disparity map of left image using our method.}
\label{fig:introduction}
\end{figure}

\begin{figure*}[t]
\begin{center}
   \includegraphics[width=17.2cm, height=7cm]{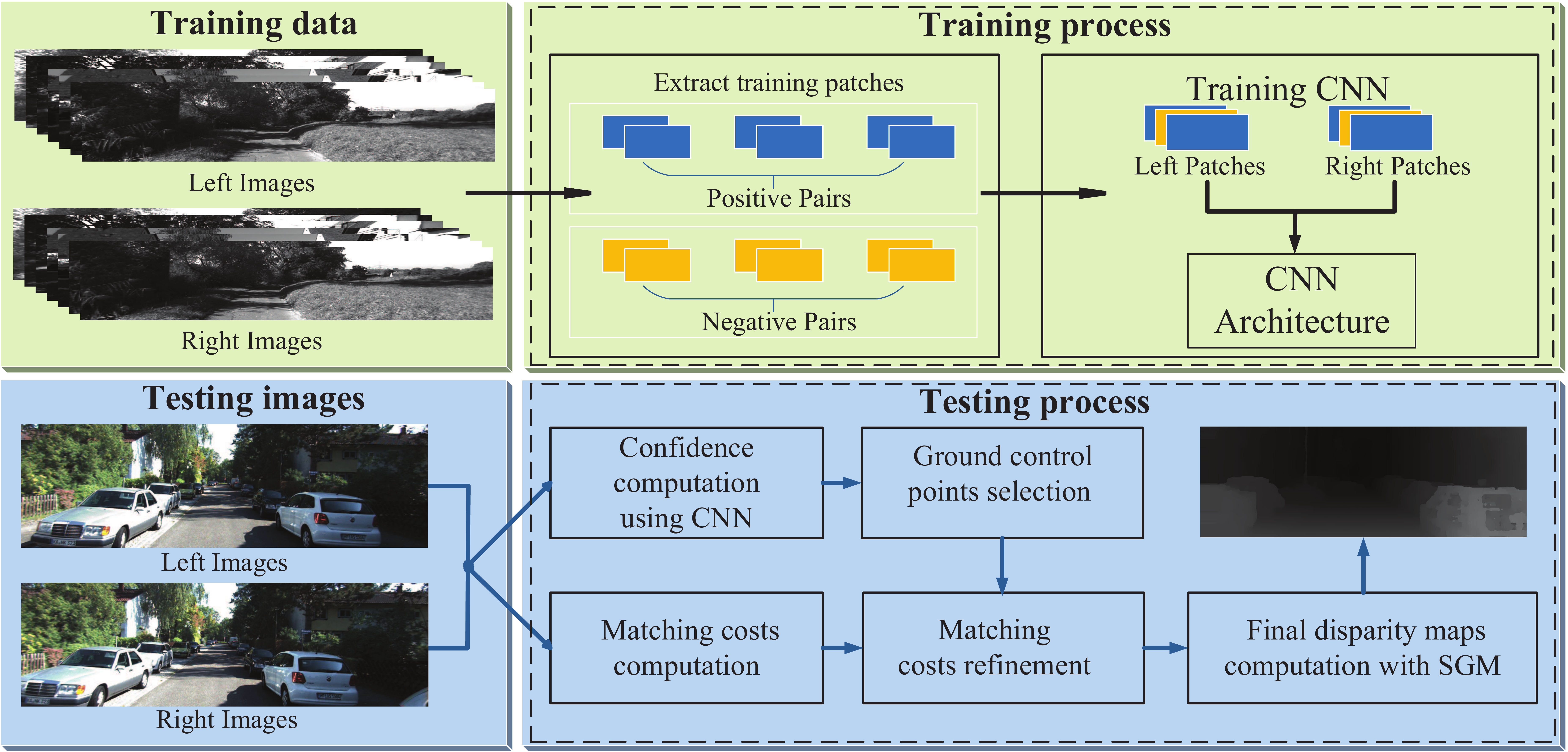}
\end{center}
   \caption{ Overview of the algorithm. The proposed algorithm is mainly composed of five stages: training CNN, confidence computation by CNN, ground control points selection, matching costs refinement, and final disparity maps computation.}
\label{fig:framework}
\end{figure*}
   According to Scharstein and Szeliski \cite{scharstein2002taxonomy}, the stereo algorithms can be classified into two main categories depending on whether the minimization procedure solving with a global cost function: Local and global algorithms. 
   In a local algorithm, the disparity computation at a given pixel depends on intensity/color values within a finite support window, and usually make implicit smoothness assumptions by aggregating pixel-based matching costs. 
   Then, an optimal disparity can be computed based on the aggregated matching costs. In comparison to global algorithms, local algorithms are generally faster, but less accurate since  narrow limitation. 
   In contrast, global algorithms make explicit smoothness assumptions and search an optimal disparity by solving an energy based optimization problem. 
   Prevalent global methods include those based on dynamic programming \cite{DYM-bobick1999large}, belief propagation \cite{bf-freeman2000learning,bf-sun2003stereo} and graph cuts \cite{g-cut-boykov2001fast}. 
   While these global algorithms have achieved impressive results, they usually require substantial computational resources. 
   Other than these two categories, a semi-global block matching stereo algorithm (SGM) \cite{SGM-hirschmuller2008stereo} was proposed. SGM is also based on a global energy cost function, but it performs optimizations along multiple directions. 
   Compared with global algorithms, SGM has reached a close accuracy with a much lower computational complexity. 
   As a result, a plenty of SGM based modified algorithms have been proposed, and have been successfully applied in the domain of stereo matching \cite{modify-sgm-hermann2013iterative,modify-sgm-spangenberg2013weighted}.
   
   Although SGM based methods have achieved noticeable results, stereo matching is still unavoidable confronted with the difficulties such as pixel indistinctiveness, depth discontinuities,  texture-less regions, and occlusions. 
   These difficulties may cause fail to compute credible matching cost between pixels or support windows. 
   Thus, although matching costs can be computed in a bad case scenario, these matching costs are not reliable enough in any location, which results in the decrease of stereo matching accuracy. 
  
   In the past few years, several papers \cite{haeusler2013ensemble,park2015leveraging,gcp-spyropoulos2014learning} have paid attention to the question whether the computed matching costs are in fact reliable. 
   Haeusler
et al. \cite{haeusler2013ensemble} proposed to learn a confidence measure from several features, and predicted the confidence by applying the random decision forest to learn a classifier. 
	Similarity, Spyropoulos and Modorhai \cite{gcp-spyropoulos2014learning} proposed  a learning-based approach to predict confidence, and leveraging the estimated confidence to select 
pretty reliable pixels as ground control points (GCPs) for improving the accuracy.
	 Park and Yoon \cite{park2015leveraging} selected effective confidence measures via regression forest, and retrained regression forest classifier to predict the confidence of a match using the selected confidence measures, as well as leveraging the predicted confidence for improving the accuracy of stereo matching.
	All of the above approaches require hand-engineering a set of features for confidence measure, and need training on the basis of specified matching costs, which means we have to train a new prediction classifier while using another matching cost computation algorithm.
	
	To overcome the problems mentioned above, in this paper, we focus on detecting the ground control points (GCPs) based on CNN, and leveraging the detected GCPs to improve the accuracy of stereo matching. 
	Figure \ref{fig:introduction} shows an example result of our algorithm.
	Unlike previous methods which used the hand-engineered features for confidence measure, our approach using convolutional neural network to detect GCPs/reliable points without designing the feature of confidence measure. Moreover, we use the detected GCPs to improve the accuracy of stereo matching with semi-global block matching (SGM). 
	Figure \ref{fig:framework} illustrates the overall flow of the proposed algorithm. 
	The contributions of this paper are summarized as follows:
		
	(1) Firstly, we train a convolutional neural network to learn the matching confidence of each pixel on a large set of pairs of small image patches where the ground truth disparities is available. Then, we detect the ground control points by the maximum confidence of each pixel over all disparities.
	 
	(2) Secondly, the stereo matching costs are refined by utilizing the confidence of the detected GCPs. Then the new matching costs are used to compute final disparity maps with semi-global matching algorithm. 
	
	(3) The experimental results on the KITTI 2012 stereo benchmark dataset show that our method significantly improves the accuracy of stereo matching overall all images. 
	
	The remainder of this paper is organized as follows. 
	In Section  2, we provide a brief overview of related work. 
	The CNN model for computing matching confidence is given in Section 3. 
	We then describe the algorithm of GCPs detection, and refinement scheme for matching costs in Section 4. 
	Experimental results and analyses are presented in Section 5, followed by the conclusion in section 6.  


\section{Related Work}
\label{section 2}
	Surveys regarding stereo methods we refer readers to the taxonomy of Scharstein and Szeliski\cite{scharstein2002taxonomy} and its companion website. 
	
	In this section, we first briefly review the learning-based methods for detecting ground control points, or predicting the confidence of matching costs. Then, the discussion of several stereo matching costs computation methods based on Convolutional Neural Network.  
	
	An early learning-based approach to stereo matching proposed by Lew et al.\cite{lew1994learning} adopted instance based learning (IBL) to select optimal feature set points for stereo matching. 
	Kong and Tao \cite{kong2004method} used nonparametric techniques to train a model to predict the probability of a potential match over three categories: correct, incorrect due to foreground over-extension, and incorrect due to other reasons.  The predicted  knowledge was then integrated into an MRF framework to improve the depth computation. 
	Later on, Kong and Tao\cite{kong2006stereo} extended their work by learning multiple experts from different normalized cross-correlation (NCC) matching windows sizes and centers, and the likelihood under each expert was then combined probabilistically into a global MRF framework for improving accuracy. 
	Motten et al. \cite{motten2012trinocular} trained a hierarchical classifier for selecting the most promising disparity with the matching costs and spatial relationship of pixels. 
	Peris et al. \cite{peris2012towards} designed a feature from cost volume, and computed the final stereo disparity using Multiclass Linear Discriminant Analysis (Multiclass LDA).

	More recently, in \cite{haeusler2013ensemble,park2015leveraging,gcp-spyropoulos2014learning}, they employed random decision forests to estimate the confidence of the stereo matching costs. 
	Haeusler et al. \cite{haeusler2013ensemble} trained a random decision forests classifier to predict the confidence of stereo matching costs by combining several confidence measures into a feature vector.
	Similarity, Spyropoulos and Modorhai \cite{gcp-spyropoulos2014learning} used a random decision forests classifier to estimate the confidence of the matching costs, and showed that the confidence information can further be used in a Markov random field framework for improving stereo matching. 
	Park and Yoon \cite{park2015leveraging} exhibited that effective confidence measures can be selected by estimating the permutation importance of each measure. In addition, they applied the confidence value to modify the initial matching cost, and inserted the new matching costs into stereo method to decrease the error of stereo matching. 
	All of these learning-based approaches have to structure a set features of confidence measures, and train a classifier based on specified matching costs, which indicates that different classifiers should be trained when applying different stereo matching cost computation methods.
	
	
	Convolutional Neural Networks (CNN) has been rapidly developed in recent years, and has been
extensively applied to deal with various computer vision tasks. 
	The most recent year, CNN has been used in stereo matching \cite{cnn-chen2015deep,cnn-zagoruyko2015learning,cnn-vzbontar2014computing,cnn-zbontar2015stereo} and achieved noticeable results. The destination of these methods is training a CNN with a large set of stereo image patches, and comparing the matching cost between image patches by the trained network. The main difference between them is the architecture of the network. On the Contrast to these methods, we train a CNN to compute matching confidence of each pixel, and detect the GCPs based on the output of CNN.

\begin{figure*}[t]
\begin{center}
   \includegraphics[width=1.0\linewidth]{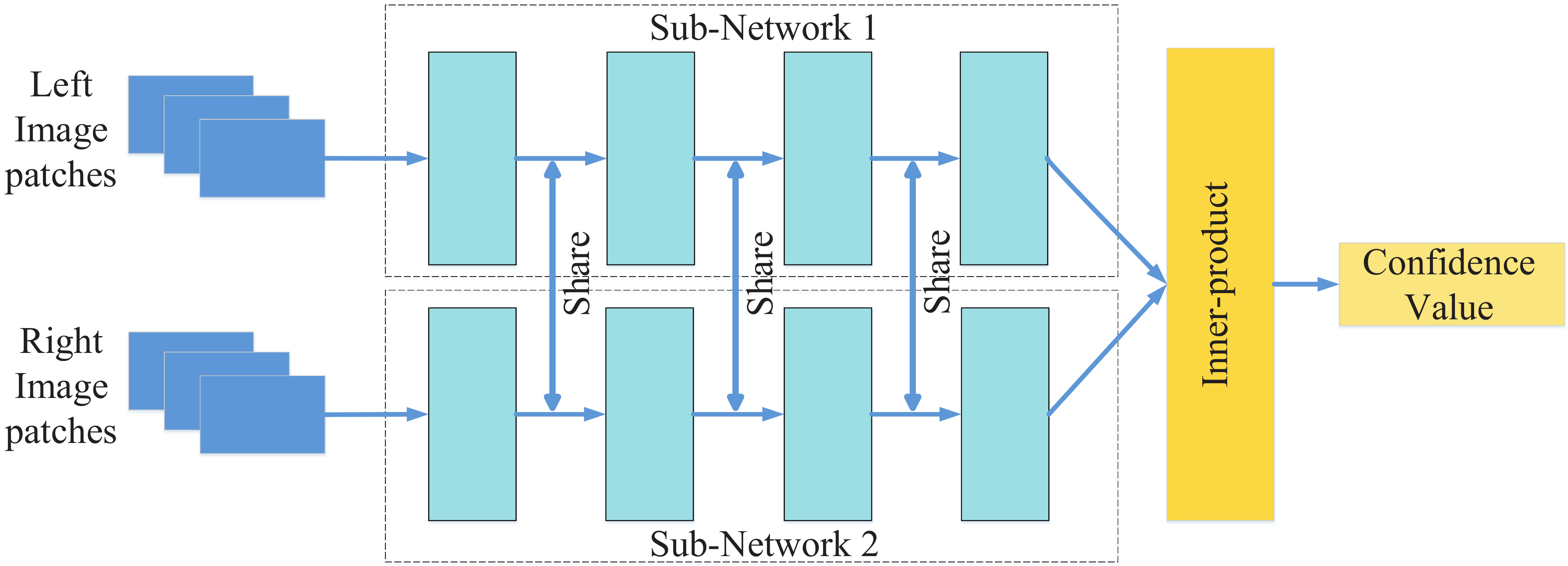}
\end{center}
   \caption{The Convolutional Neural Network architecture of our training model. Each sub-network contains four convolutional layers followed by ReLu. The confidence value is an inner product of two extracted feature vectors.}
\label{fig:CNN-structure}
\end{figure*}

\section{CNN based GCP Detection}
	In this section, we describe the details of CNN model for measuring the matching confidence of each pixel.

\subsection{Matching confidence}

 Given a pair of rectified stereo images, \(I^L\) and \(I^R\), a typical stereo algorithm begins by computing a matching cost at each position \(p\) with different disparities $d\in\left [0, d_{max} \right]$.
  Thus, a similarity/matching cost between two pixels can be compute by:
  \begin{center}
    \begin{eqnarray}
   C(p,d)=f(I^L(p),I^R(p-d))
\end{eqnarray}
  \end{center}

where $I^L(p)$ and $I^R(p-d)$ denote the patches centered at $p^L=(x,y)$ in left image and $p_d^R=(x-d,y)$ in right image, respectively. $f(\cdot)$ is the matching cost computation method.

  In a real-world scene application of stereo matching, we may suffer from the difficulties such as depth discontinuities,  texture-less regions, and occlusions. These bad scenarios may generate unreliable matching costs, which degrades the overall performance of stereo matching. Thus, if we can estimate the matching confidence that reveal the reliability of matching cost, the stereo matching performance will be promoted with the confidence information.
  To address this issue, we attempt to use CNN-based model to calculate the matching confidence between patches, where the two patches are centered at the same 3D points having a high matching confidence, and low when they are not.
   

\subsection{Training Datasets}
	In order to compute the matching confidence between patches, the input to the neural network is a pair of image patches, and the output is a inner-product that measures the matching confidence between the inputs.
	
	Hence, a training example is composed of two image patches, one from left image,  $p^L=(x,y)$, and one from right image,  $p_d^R=(x-d,y)$. We sample one negative and one positive example for each location where the true disparity $d_t$ is known.
	
	A negative example is obtained by setting the center of the right patch to
	
	\begin{center}
	$p_d^R = (x-d_t+o_{neg},y)$
	\end{center}
where $o_{neg}$ is a random offset from the interval [$-N_{high}, ..., \\ -N_{low},
	N_{low}, ... ,N_{high}$]. This random offset guarantee that the image patches of negative samples are not centered at the same 3D point.
	
	Similarity, a positive example is obtained by setting
	
	\begin{center}
	$p_d^R = (x-d_t+o_{pos},y)$
	\end{center}
	where $o_{pos}$ is a random offset from the interval [$-P_{high},...,\\ P_{high}$].

\subsection{Architecture}
 
  An overview of our architecture is shown in Figure \ref{fig:CNN-structure}. This architecture is a siamese network. There are two sub-networks in the network that share exactly the same architecture and the same
set of weights. The input is a pair of $9 \times 9$ grayscale image patches, centered at $p^L=(x,y)$ and $p_d^R=(x-d,y)$. Each sub-network takes one of the two patches as input, and it is a composition of a number of convolutional layers with rectified linear units (ReLU) layers following all  convolutional layers, we do not use 
max-pooling layer to preserve spatial-variance. For a pair of patches, we apply a 4-layer CNN model to independently extract feature descriptors $F(I)$ from each sub-network to represent each input patch, and use the inner-product layer to calculate the matching confidence between two  descriptors $F(I^L)$ and $F(I^R)$. The size of each convolution kernel is $3 \times  3$, and the number of feature maps in each layer is $64$.

	We train our model by minimizing a hinge-based loss term which is employed in \cite{cnn-zbontar2015stereo}. The hinge-baed loss term is defined as:
	\begin{eqnarray}
	\max(0,\epsilon+ s_{neg} - s_{pos})
	\label{equation:max}
\end{eqnarray}

	This loss term is computed by considering pairs of training examples centered at the same image position, with one example which $p^L=(x,y)$ corresponds to $p_d^R=(x-d,y)$ as positive, and one which not corresponded as negative. $s_{pos}$ denotes the output of the network for the positive example, $s_{neg}$ denotes the output of the network for the negative example, and $\epsilon$ is a positive real number that we set it to 0.2 in this paper.
	
	In the training process, we input a set of examples over all example pairs, and compute the loss by summing the terms in Equation \ref{equation:max}. Thus, we can compute the matching confidence for each pair of patches by using the trained CNN model.
	We normalize the matching confidence into the interval $\left [0, 1\right ]$.

\section{Improving Stereo Matching With Ground Control Points}
	In this section, we propose our ground control points (GCPs) detection approach  based on the matching confidence computed by CNN model, and refine the matching costs depending on the confidence of GCPs.
	
	
\subsection{Detecting ground control points}	
	According to the previous definitions, a GCP is defined as a pixel with a high confidence that the computed matching costs are reliable.
	
	Given a pair of stereo images, a matching confidence volume, $Vol(p,d)$, could be obtained by performing the forward pass for each pixel and each disparity under consideration. For each pixel $p$ in the left image, the maximum matching confidence, $Cof_c(p)$, is computed by maximizing the matching confidence volume in each disparity:
	
	\begin{eqnarray}
	{Cof}_c(p) = {\underset{d}{\max}} \; Vol(p,d)
	\end{eqnarray}
The maximum matching confidence indicates the reliability of matching costs for each pixel. We also compute the most confident disparity, $Cof_d(p)$, for each pixel $p$:
		\begin{eqnarray}
	Cof_d(p) = {\underset{d}{\arg\max}} \; Vol(p,d)
	\end{eqnarray}
	
	We propose a simple manner to select GCPs using the maximum matching confidence of each pixel, the selected GCPs can be used to impact neighboring pixels. The selecting criterion is followed by: if ${Cof}_c(p)$ is larger than a constant threshold $\theta$, pixel $p$ is a GCP, otherwise not (i.e. unreliable pixels).	We define a pixel $p$ as $p_{GCP}^+$ if it belongs to GCP, otherwise, define it as $p_{GCP}^-$.
	
	The main challenge in GCPs selection step is the trade-off between density and accuracy.
	Considering that too much GCPs are selected, some of them may contain unreliable matching costs that will be propagated to neighboring pixels, which may cause the decreasing of the stereo matching accuracy. Conversely, few GCPs will be not effective enough to improve the matching performance.   	
	
	Therefore, aiming at selecting the GCPs that are able to improve the accuracy of stereo matching, we learn the threshold on the maximum matching confidence $Cof_c(p)$ by cross-validation. 
	
\subsection{Refining matching costs with GCPs}
	In the previous subsection, we described an approach to select GCPs using maximum matching confidence. In this subsection, we present a matching costs refinement approach using the selected GCPs that can be used for the final optimization.
		
	Given a pair of stereo images, we can compute a matching costs volume, $C(p,d)$, by a matching cost computation method, such as, normalized cross-correlation (NCC), sum of absolute differences (SAD), and census-based Hamming distance (Census). 
	The refinement scheme is a two-step process.
	In the first step, based on the observation that $p_{GCP}^-$ contains unreliable matching costs, we refine the matching costs of $p_{GCP}^-$  by setting the cost of all disparity to a constant high cost value $C_{GCP}^{hi}$. As a result, the wrong matching cost could be avoided to pollute other reliable pixels. 
	In the second step, we refine the matching costs of the selected $p_{GCP}^+$, by setting the matching cost to a constant low cost value $C_{GCP}^{low}$ for the most confident disparity of $p_{GCP}^+$. The most confident disparity of $p_{GCP}^+$ is $Cof_d(p)$, that was computed in the last subsection. The matching costs of all the other disparities for $p_{GCP}^+$ are unmodified. 
In this way, we the matching costs are refined. We define the new  matching costs volume as $\hat{C}(p,d)$. 
	Therefore, unreliable pixels can be easily dominated by more confident pixels in the subsequent optimization process, and the more confident disparities can obtain a lower value,
 	so that reduce the disturbance from other unreliable disparities. 
 	Moreover, it is worth noting that the proposed matching cost refinement scheme can be used for any matching cost computation algorithms.
 
 In order to demonstrate our approach is robust to various matching cost computation algorithms, we compute two different cost volumes, sum of absolute differences (SAD) and census-based Hamming distance (Census), respectively. 
 
 The SAD matching costs is calculated as follows:
 		 \begin{eqnarray}
 \begin{array}{l}

	C_{SAD}(p,d) = \underset{q\in N(p)}{\sum} \left |I^L(q)-I^R(qd)  \right |
	\end{array}
\end{eqnarray}
where $I^L(q)$ and $I^R(qd)$ are intensities of pixel at $q=(x,y)$ in the left image, and pixel at $qd = (x-d,y)$ in the right image. And $N(p)$ is the set of pixels within a fixed support window centered at $p$. 

 And the Census is defined as follows:
 
 \begin{eqnarray}
 \begin{array}{l}

	C_{Census}(p,d)=\underset{q\in N(p)}{\sum} 
	XOR(W^L(p,q),W^R(pd,qd))
	\end{array}
\end{eqnarray}
where $pd$ is the pixel at position $pd = (x-d,y)$ in the right image corresponding to $p$ in the left image, and $W(p,q)$ is a binary transform function. If the intensity of $p$ is larger than $q$, $W(p,q)$ returns 1,  
 otherwise zero. $N(p)$ is the set of pixels within a fixed support window centered at $p$.
\subsection{New matching cost with SGM}
 We compute the disparity map by applying the refined cost volume $\hat{C}(p,d)$ to the semi-global matching (SGM) \cite{SGM-hirschmuller2008stereo} algorithm. Following Hirschmuller \cite{SGM-hirschmuller2008stereo}, the SGM considers the stereo matching as an energy function that minimizes:
 	
 \begin{eqnarray}
 \begin{array}{l}

	\displaystyle E(D) = \underset{p}{\sum}(\hat{C}(p,D_p)+ 
	\underset{q\in N(p)}{\sum}P_1 \cdot  1\left [\left |D_p-D_q  \right |=1 \right ]\\
	\displaystyle \qquad \qquad + \underset{q\in N(p)}{\sum}P_2 \cdot  1\left [\left |D_p-D_q  \right |>1 \right ])
	\end{array}
	\label{equation-sgm}
\end{eqnarray}
where $1\left [ \cdot \right ]$ denotes the indicator function. 
	The first term is the sum the refined matching costs of all pixels, so that penalizes disparities $D_p$ with high matching costs.
	The second term adds a small constant penalty $P_1$ for all pixels $q$ having small disparity differences in the neighborhood $N_p$ of $p$.  
	The third term adds a larger constant penalty $P_2$ for all pixels $q$ having disparity differences larger than 1 in the neighborhood $N_p$ of $p$. The minimization of  the equation \ref{equation-sgm} is an NP-hard problem, instead of performing minimizing $E(D)$ in all directions simultaneously, we perform the minimization in a single direction, and repeat for 16 directions. The cost $L_r(p,d)$ along a path in the direction $r$ of the pixel $p$ at disparity $d$ is defined recursively as:
 \begin{eqnarray}
 \begin{array}{l}
 
L_r(p,d) = \hat{C}(p,d) + \min(L_r(p-r,d)),\\
\qquad	\qquad	\qquad \qquad	L_r(p-r,d-1) +P_1,\\
\qquad	\qquad	\qquad \qquad	L_r(p-r,d+1) +P_1,\\
\qquad	\qquad	\qquad \qquad	\underset{k}{\min} L_R(p-r,k)+P_2)

	\end{array}
\end{eqnarray}

The final disparity costs could be obtained by averaging the costs along 16 directions:

    \begin{eqnarray}
 \begin{array}{l}
 
L(p,d) = \frac{1}{16}\underset{r}{\sum}L_r(p,d)

	\end{array}
\end{eqnarray}   

Finally, the disparity image $D(p)$ is computed using the Winner-Takes-All (WTA) strategy as follows:

    \begin{eqnarray}
 \begin{array}{l}
 
D(p) = \underset{d}{\arg\min}L(p,d),\quad  d\in\left [0, d_{max}  \right ]

	\end{array}
\end{eqnarray}  
where $d_{max}$ is the maximum value of possible disparities.

	 \begin{figure}[t]
\begin{center}
   \includegraphics[width=1.0\linewidth, height = 0.7 \linewidth]{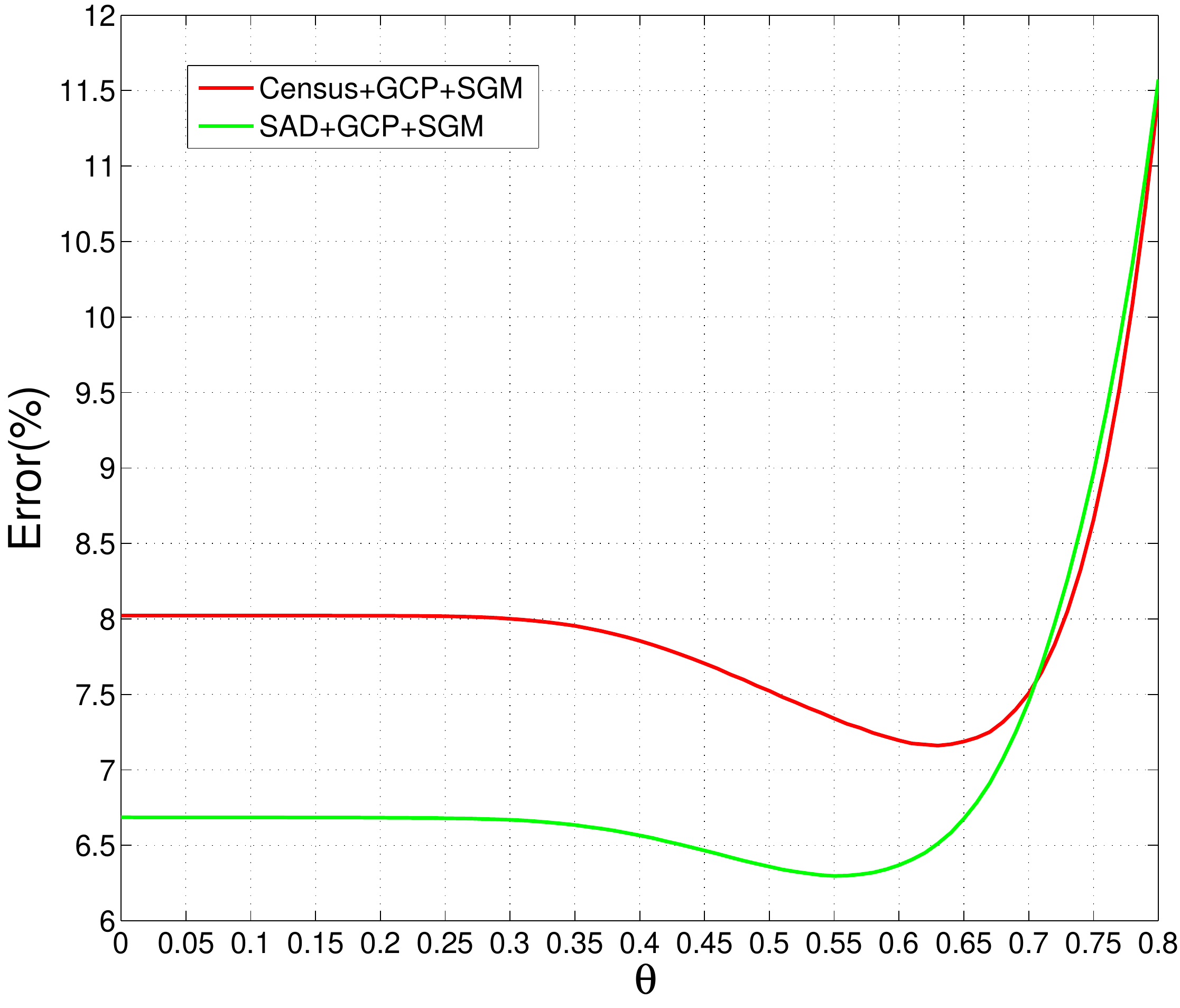}
\end{center}
   \caption{Evaluate the performance under different confidence parameter settings of $\theta$. We fix the parameters  $\left \{  C_{GCP}^{hi} = 200 ,C_{GCP}^{low} = 1.3 \right \}$ for Census, and $\left \{  C_{GCP}^{hi} = 5 ,C_{GCP}^{low} = 0.001 \right \}$ for SAD, respectively.}
\label{fig:threshold}
\end{figure}

\section{Experiments}
	In this Section, we evaluate the performance of the proposed method on the KITTI 2012 stereo benchmark dataset, and compare with the state-of-the-art learning-based method (Lev) \cite{park2015leveraging}. We use Torch7 \cite{collobert2011torch7} to train our CNN, and the stereo method is implemented in CUDA, with a Nvidia GeForce GTX Titan GPU.
	
\subsection{Data set}
	The KITTI 2012 stereo benchmark dataset\cite{geiger2013vision} is composed of 194 training and 195 testing high-resolution images, which are collected from two video cameras on an autonomous driving platform around a urban environment. 
The goal of the KITTI stereo benchmark dataset is to estimate the true disparities for all pixels on the left image. 
	 The ground truth disparities for the training images are public provided for researchers, and the one for testing images are withheld, researchers can evaluate their method on the test set by submission the result to the website.

\subsection{Parameter settings}
	To compute pixel-wise matching costs, we set the support window size to $9 \times  9$ for both SAD and Census. For the penalty teams in the SGM, we set $\left \{  P_1 = 1 ,P_2 = 14 \right \}$ for SAD, and $\left \{  P_1 = 4 ,P_2 = 128 \right \}$ for Census, respectively.  The training parameters of CNN are set to \{$   N_{high} = 8 , N_{low} = 4,  P_{high} = 1 $\}. We fix the parameters  \{$ C_{GCP}^{hi} = 200 ,C_{GCP}^{low} \\ = 1.3 $\} for Census, and $\left \{  C_{GCP}^{hi} = 5 ,C_{GCP}^{low}  = 0.001 \right \}$ for SAD, respectively. The matching costs are ranged in $\left [0, 80 \right] $ for Census, and $\left [0, 3.2 \right]$ for SAD,respectively.
	
	We train the CNN using stochastic gradient descent to minimize the hinge-based loss. We preprocess each image by subtracting the mean and dividing by the standard deviation of its pixel intensity values.
	
\subsection{Confidence parameter analysis}
	The goal here is to evaluate the performance under different confidence parameter settings of $\theta$.
	Results evaluated on KITTI training dataset are summarized in Figure \ref{fig:threshold}. It can be seen that the best result is obtained nearby $\theta = 0.60$.
	Setting too large value of $\theta$ would not achieve improving results even bring about decrease sharply.
	Setting too large value of $\theta$ may consider a major of pixels as unreliable points, result in losing the information of reliable pixels, which does not help to improve the matching accuracy even harm the matching performance.
	A too low value may consider most of the pixels to GCPs, this may only use the information of CNN matching confidence, and fail to avoid the influence of unreliable pixels. 
	  We set $\theta = 0.55$ for SAD, and $\theta = 0.60$ for Census in all following experiments.

%

\begin{figure*}[t]
\begin{center}
   \includegraphics[width=1\linewidth]{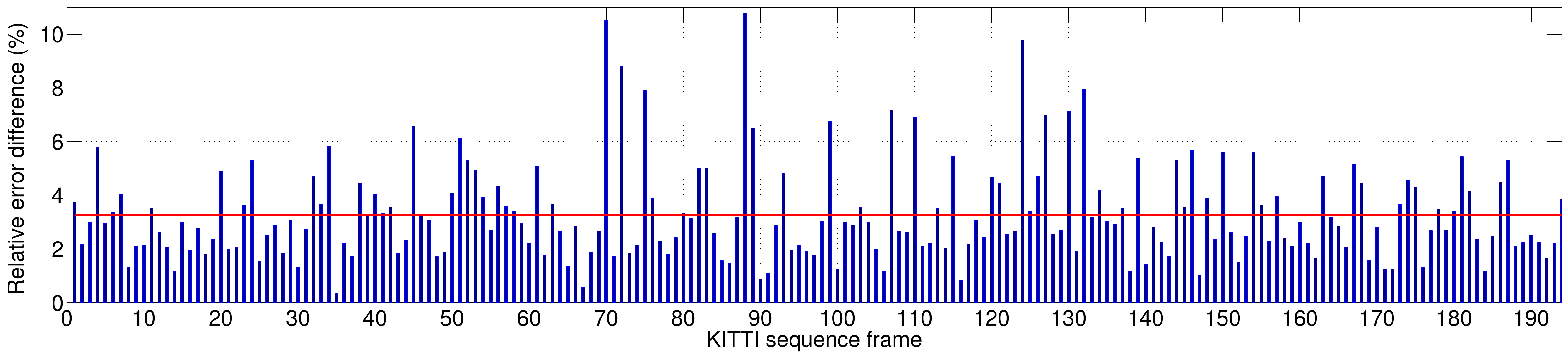}
\end{center}
   \caption{Overall improvement compared Census+GCP+SGM with Census+SGM algorithm. Each bar indicates the performance improvement of the proposed algorithm using Census matching costs, and horizontal line indicates average improvement in terms of the error rate. Our algorithm outperforms the census+SGM algorithm in all frames.}
\label{fig:overall-census}
\end{figure*}

\begin{figure*}[t]
\begin{center}
   \includegraphics[width=1\linewidth]{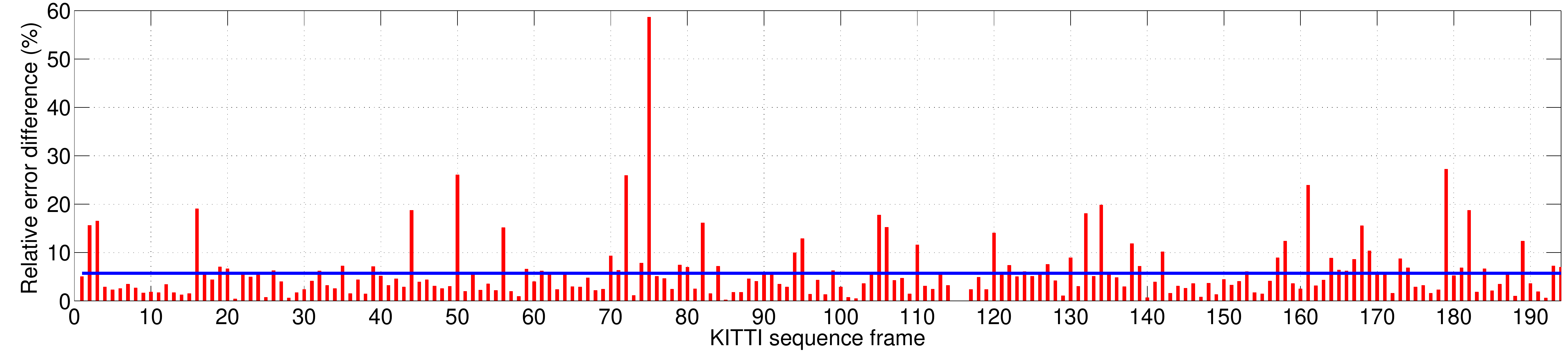}
\end{center}
   \caption{Overall improvement compared SAD+GCP+SGM with SAD+SGM algorithm. Each bar indicates the performance improvement of the proposed algorithm using SAD matching costs, and horizontal line indicates average improvement in terms of the error rate. Our algorithm outperforms the SAD+SGM algorithm in all frames.}
\label{fig:overall-ad}
\end{figure*}

\begin{table}
\begin{center}
\begin{tabular}{l|c}
\hline
\hline
Method & Error \\
\hline

Census+SGM & 10.46\% \\
SAD+SGM & 12.04\% \\
Census+GCP+SGM &  7.19\%\\
SAD+GCP+SGM & 6.29\% \\
\hline
\hline
\end{tabular}
\end{center}
\caption{Error rates of the final disparity maps after SGM using different matching cost methods. 
Our method (+GCP) has a much lower error rate than that of without GCP refining.}
\label{tabel:result1}
\end{table}

\subsection{Stereo performance improvement analysis}
	We first evaluate the performance improvement of our approach while applying different matching costs computation methods, SAD and Census. The results evaluated on KITTI training dataset are listed in Table \ref{tabel:result1}.

\begin{figure*}[t]
\begin{center}
   \includegraphics[width=1\linewidth]{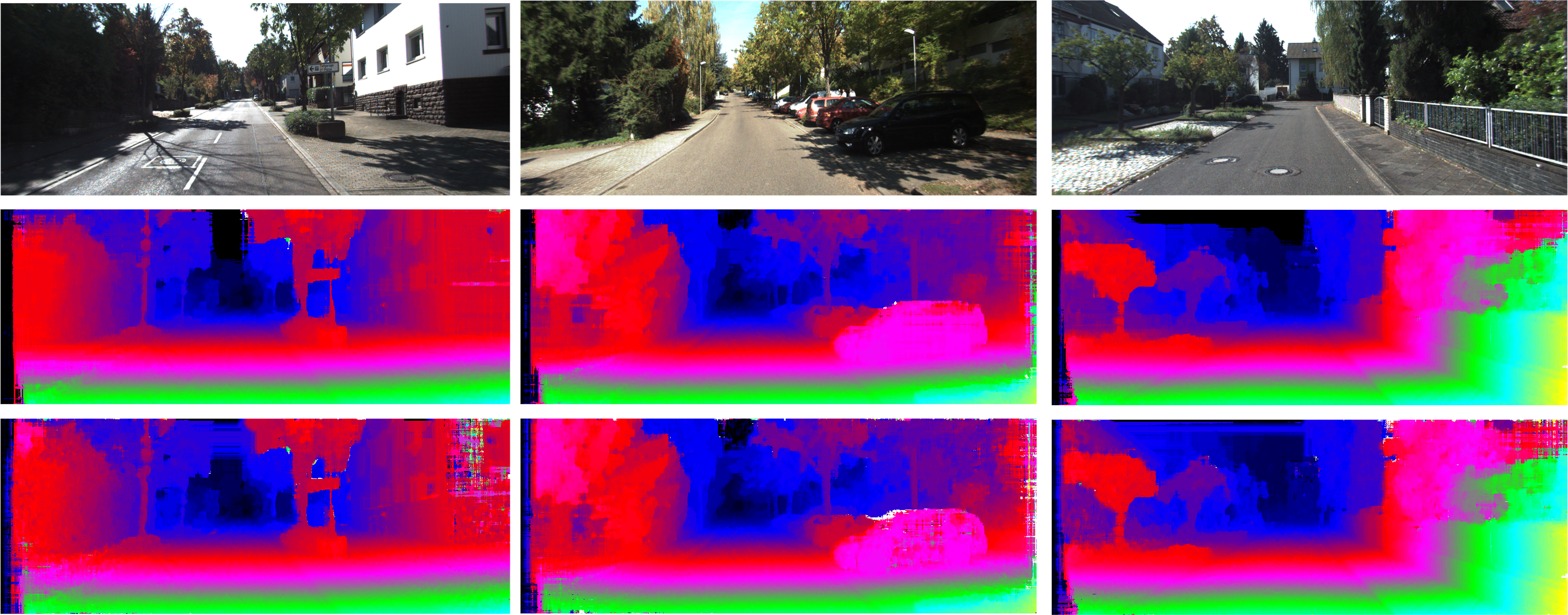}
\end{center}
   \caption{Examples of disparity maps on the KITTI 2012 data set. \textbf{From top to bottom:} left images; 
   disparity maps of SAD+GCP+SGM;
  and disparity maps of Census+GCP+SGM.
   }
\label{fig:compare}
\end{figure*}

	The proposed method using the new matching costs that are refined by GCPs significantly improves the accuracy of disparity maps under different matching cost computation methods. The average error rate of the method which uses the original Census matching costs (Census+SGM) is $10.46\%$\\, while the modified method (Census+GCP+SGM) which uses refined matching costs reduces the error rate by $3.26\%$. 
	Similarly the error rate of SAD+GCP+SGM is significantly reduced  by $5.75\%$.
	The overall improvement is detailed in Figure \ref{fig:overall-census} and Figure \ref{fig:overall-ad}.
	We see that the proposed approach consistently improves the accuracy of stereo matching over all images. In addition, our approach is robust to different matching cost  computation methods.	
	Figure \ref{fig:compare} contains some examples for the disparity maps produced by our method.

	Secondly, we compare our approach against the state-of-the-art learning-based method \cite{park2015leveraging}. 
	The results evaluated on KITTI training dataset is shown in Table \ref{tabel:result2},
	the proposed method achieves an error rate of 7.19\% which is lower than that of the lev\cite{park2015leveraging} by 2.19\%.
	It demonstrates that our method obtains more reliable GCPs, and makes full use of the confidence information of GCPs for improving the accuracy of stereo matching. Specifically, our proposed method could directly be applied to different matching costs computaion methods, while Lev\cite{park2015leveraging} requires to train a new predictor for another matching costs.
	 
	
	Note that, in order to observably reflect the improvement of our method, we do not apply any post-processing algorithms, such as left-right consistency checking and subpixel enhancement, in these experiments.

\begin{table}
\begin{center}
\begin{tabular}{l|c|c|c}
\hline
\hline
Method & Census+SGM & Our & Lev\cite{park2015leveraging}\\
\hline

Error &  10.46\% & 7.19\% & 9.38\% \\
\hline
\hline
\end{tabular}
\end{center}
\caption{Error rates of the final disparity maps comparison.
The initial matching costs are computed by Census. }
\label{tabel:result2}
\end{table}

%

\section{Conclusion}
	This paper presents a Convolutional Neural Network based approach that is able to detect the ground control points (GCPs) according to the matching confidence of each pixel. We first learn a Convolutional Neural Network to estimate the confidence of each pixel. Then we select GCPs of image depending on the confidence. In addition, we present a robust approach to obtain a new matching costs by refining the matching costs with the GCPs confidence, which can further be used to compute the final disparity maps. Experiments on KITTI 2012 stereo dataset demonstrate that our approach  significantly improves the accuracy of stereo matching on overall images, and our approach achieves an  impressive result that surpasses the current leading learning-based method. Furthermore, our proposed method can be applied in various matching cost computation methods.

\begin{acknowledgements}
This work is supported by the Nature Science Foundation of China (No.61202143, No. 61572409), the Natural Science Foundation of Fujian Province (No.2013J05100) and Fujian Provi-nce 2011 Collaborative Innovation Center of TCM Health Management.
\end{acknowledgements}


\bibliographystyle{spmpsci}
\bibliography{egbib}




\end{document}